# Persian Speech Emotion Recognition by Fine-Tuning Transformers

Minoo Shayaninasab, Bagher Babaali*

**Abstract**

Given the significance of speech emotion recognition, numerous methods have been developed in recent years to create effective and efficient systems in this domain. One of these methods involves the use of pretrained transformers, fine-tuned to address this specific problem, resulting in high accuracy. Despite extensive discussions and global-scale efforts to enhance these systems, the application of this innovative and effective approach has received less attention in the context of Persian speech emotion recognition. In this article, we review the field of speech emotion recognition and its background, with an emphasis on the importance of employing transformers in this context. We present two models, one based on spectrograms and the other on the audio itself, fine-tuned using the shEMO dataset. These models significantly enhance the accuracy of previous systems, increasing it from approximately 65% to 80% on the mentioned dataset. Subsequently, to investigate the effect of multilinguality on the fine-tuning process, these same models are fine-tuned twice. First, they are fine-tuned using the English IEMOCAP dataset, and then they are fine-tuned with the Persian shEMO dataset. This results in an improved accuracy of 82% for the Persian emotion recognition system.

## 1. Introduction

Emotion is a psychophysical process that plays a crucial role in human decision-making, interaction, and cognitive processes. With advancements in our comprehension of emotions and their functions, there is a growing demand for automatic emotion recognition systems. These systems are extensively studied and applied using speech, text, facial cues, and physiological signals, either individually or through multimodal approaches. As everyday intelligent devices proliferate, one of the challenges in this field is to make interactions with these devices more human-centric, moving away from traditional machine-human interactions [1].

With the growth of audio and image datasets in the past decade, developing systems for emotion recognition from text, speech, and images and fusing them has become a significant research area in human-computer interactions. Among unimodal studies, speech as a modality is particularly promising, given the long history of audio processing, the availability of suitable hardware and software infrastructure [3].

Research has shown that, in human communication, 7%, 38%, and 55% of information is conveyed through text (content of speech), voice, and facial expressions (images), respectively [2]. Notably, there is a significant sensory load distinction between speech and its text form, creating an opportunity for focusing on speech-based emotion recognition systems. Developing such systems is intriguing because most studies and models focused on speech often emphasize the textual content and speaker-related features, neglecting information that conveys emotions regardless of words and speakers, considering it as additional, and even intrusive.

In the Persian language, due to a lack of diverse datasets, less research has been conducted on speech emotion recognition. Recent years have seen some progress, but there is still much room for research and development.

The diversity of modeling tools in artificial intelligence and the vast digital data available in recent years allow the pursuit of emotion recognition systems through various methods. One of these methods is deep learning, specifically using self-supervised learning with large unlabeled datasets to train models that can later be used for feature extraction and fine-tuning in various tasks. These self-supervised models were initially used for text and achieved State of the Art performance using the transformer architecture. This architecture was quickly adapted for various data types, especially in audio and speech processing, as common and robust tools [3].

In this article, after reviewing research on speech emotion recognition, self-supervised models using transfer learning are employed to construct a Persian speech emotion recognition system. Two models, one based on spectrograms and the other on speech audio, will be fine-tuned and compared with prior work. Then, these models will be fine-tuned once with emotional speech from the IEMOCAP dataset in English and again with Persian emotional speech. The impact of cross-lingual learning will be investigate using these results.

## 2. Related work

Approaches to speech emotion recognition involve two stages: feature extraction and classification. In the first stage of speech processing, features such as source-based arousal features, prosodic features, and vocal tract articulations are extracted. In the second stage, classifiers like Support Vector Machines or neural networks are used for the classification of the extracted features [10].

Recent unimodal studies on speech emotion recognition have focused on identifying relevant audio features, such as fundamental frequency (pitch), speech intensity, bandwidth, and duration. Prior to the widespread use of deep learning, speech emotion recognition employed methods like Hidden Markov Models, Gaussian Mixture Models, and Support Vector Machines. These approaches required significant feature engineering, and any changes in features often necessitated the reconstruction of the entire system. The advent of deep learning in this field significantly increased the accuracy of results in controlled environments, raising it from around 70% to over 90% [4].

The diversity of extracted speech features, as well as the variety of tools and architectures used, follows the field of speech emotion recognition. Recurrent neural networks, especially bidirectional long short-term memory networks, have demonstrated good performance due to their ability to model temporal features. This approach employs direct speech and acoustic features.

Le et. al [9] introduced the idea of using uncertainty in emotional labels and an efficient learning algorithm. They employed a bidirectional long short-term memory (BiLSTM) recurrent network architecture and reported a weighted accuracy of 62.85 on the IEMOCAP dataset. This number represents a 12% improvement compared to the baseline by extreme learning machines.

Another architecture in this field involves Convolutional Neural Networks (CNNs). CNN-LSTMs have gained recent attention. This approach utilizes the audio spectrogram and its visual features. Satt et. al [12] presented a new implementation of speech emotion recognition. They segmented speech into smaller units of less than 3 seconds and employed the audio spectrogram as features. Ultimately, they improved the performance of the CNN network from 64 to 68 on the IEMOCAP dataset by incorporating long short-term memory.

One of the recent developments in the field of Persian speech emotion recognition is the work of Yezdani [10], in which deep learning models, including LSTM-CNN and LSTM-RNN with and without attention mechanisms, were applied to low-level and high-level signal features from the emotional speech dataset shEMO. This research reported an unweighted accuracy of 65.2 on this dataset, whereas the baseline accuracy using Support Vector Machines was reported as 58.2 in the paper introducing the dataset [11].

In recent years, self-supervised architectures like HuBert, Wav2vec2.0, and Whisper have shown promising results in speech emotion recognition. Kakouris et. al [5], through fine-tuning WavLM on the IEMOCAP dataset, reported an accuracy of 75, which has been the best performance so far.

Self-supervised methods for speech emotion recognition provides room for further research. Numerous experiments have been undertaken to enhance the robustness of emotion recognition systems, covering various aspects such as speech features, speaker-related features, and architectural choices. The fusion of speech features with text or facial expressions has been explored to improve system accuracy, referred to as multi-modal emotion recognition [6-8]. One of the advantages of these architectures is the ability to fine-tune pre-trained models for a new language, which provides flexibility for multilingual or cross-lingual models. However, the incorporation of pre-trained self-supervised models into Persian language emotion recognition has not been implemented to date.

### 3. Dataset and transformer models

One of the large-scale datasets for Persian emotion recognition is the shEMO emotional speech dataset, introduced in 2020 [11]. shEMO comprises 3,000 semi-natural utterances, equivalent to 3 hours and 25 minutes of speech data extracted from online radio scripts. The labels cover five basic emotions, including anger, fear, joy, sadness, and surprise, as well as a neutral state. The labeling was performed by 12 separate annotators, and the final labels were determined based on the aggregation of their judgments. There are a total of 87 Persian speakers, including 31 females and 56 males, reflecting a gender imbalance among the speakers. The duration of the utterances varies between 0.35 to 33.32 seconds, with an average of 4.11 and a standard deviation of 3.41 reported. The highest average is observed in the neutral class, and the lowest in the surprise class.

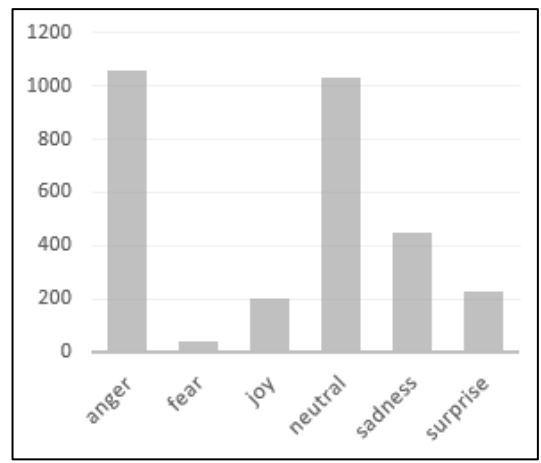

**Figure 1:** Label Frequencies in the shEMO Dataset

In terms of label frequency, there is a significant imbalance in the data (Figure 1). All six labels participated in the following experiments, despite the general imbalance among labels and the insufficient data for the fear class.

The first model used is the base model of wav2vec2.0 [13]. This powerful model is built on a transformer architecture with a self-attention. It has undergone pretraining on 960 hours of 16 kHz audio and is accessible for various tasks in open-source form. It directly accepts raw audio as input, eliminating the necessity for separate feature extraction methods.

The second model used is AST (Audio Spectrogram Transformer), which was introduced for the first time in the research [14]. This model is essentially a vision transformer [15]. The vision transformer eliminated the necessity for using convolutional networks for image data. The audio spectrogram transformer has a similar architecture, and it has been pretrained on spectrogram images, making it available open-source for fine-tuning purposes. This model has demonstrated superior performance compared to other models on many sound classification tasks.

### 4. Experiments and results

The 3000 speech segments, as described earlier, are randomly divided into training, validation, and test sets in a ratio of 80%, 10%, and 10%, respectively. This categorization remains consistent throughout all experiments. Feature extraction, preprocessing, and data representation are entirely based on the mentioned models, and the duration of speech input is set to 5 seconds. In the first part of the experiment, the two introduced transformer models are fine-tuned on the training set. The accuracy of each model on the test data is reported in Table 1.

**Table 1:** Performance of Fine-Tuned Models

| model | accuracy |
|---|---|
| Fine-tuned Wav2vec2.0 | 80 |
| Fine-tuned AST | 79.3 |

In the second part of the experiment, two models are employed, which are the results of modifying the previously introduced models. These two models have been fine-tuned on IEMOCAP speech data for 4-class emotion recognition. By changing the last layer from 4 to 6 classes, they are fine-tuned once again on the shEMO data as in the previous section. The accuracy on the test set is reported before and after this fine-tuning (Table 2).

**Table 2** Accuracy of fine-tuned models on IEMOCAP data before and after re-fine-tuning on shEMO

| model | Accuracy before fine-tuning on shEMO | Accuracy after fine-tuning on shEMO |
|---|---|---|
| Finetuned-Wav2vec2.0 | 25.3 | 82 |
| Finetuned-AST | 25 | 81.7 |

### 5. Discussion

In introducing the shEMO dataset for emotion recognition, machine learning methods were used. In this classification, the fear emotion group was entirely removed from the classification problem due to insufficient data. Over the remaining 5 groups, a Support Vector Machine (SVM) model achieved an average accuracy of 58.2 [11]. In [10], a more recent classification using audio features and neural networks improved this accuracy to 65.2. In the first experiment in this article, both models, regardless of their working method and the features they extract, significantly outperformed the results mentioned earlier. Although the results in Table 1 are not averaged over different data splits, the notable difference in performance demonstrates the capabilities of these two models. The wav2vec2.0 model has performed slightly better.

The second experiment answers the question of whether fine-tuning on speech from another language and then fine-tuning with Persian data can improve the final model's accuracy or not. Both of these models perform better before learning Persian speech from random initialization. After fine-tuning on shEMO, the accuracy for the wav2vec2.0 model improves by

2%, and for the AST model, it improves by 2.4%. Rows in Table 1 and 2 are directly comparable because the training, model loading, evaluation, and test accuracy measurement processes for each model in the second experiment are exactly the same as in the first experiment. Fine-tuning with IEMOCAP data was also done for both models with the same training and evaluation datasets. Since the increase in accuracy is evident for both models in both rows of Table 2 compared to Table 1, it can be concluded that there is a cross-lingual relationship among emotional speech features.

### 6. Conclusion

In this paper, the best accuracy for Persian speech emotion recognition has been achieved using transformers compared to previous methods. The superiority of self-supervised pre-trained models for this task has also been demonstrated. Furthermore, given the positive impact of emotional data from other languages on the performance of, there is room for improving Persian speech recognition models, even though diverse datasets for Persian language is insufficient. It is important to continue developing rich datasets for Persian speech.

The best model developed in this paper, which is currently used for Persian speech emotion recognition, has an accuracy of 82% on shEMO. However, due to the close performance of the two final models presented, cross-validation is necessary to select the superior model, which would require more time and computational resources and will be considered in the future. For practical applications, it's essential to report and compare other metrics such as weighted accuracy and f-score in addition to accuracy.